\documentclass[sigconf]{acmart}
\AtBeginDocument{%
  \providecommand\BibTeX{{%
    \normalfont B\kern-0.5em{\scshape i\kern-0.25em b}\kern-0.8em\TeX}}}

\setcopyright{acmcopyright}
\copyrightyear{2021}
\acmYear{2021}
\acmDOI{10.1145/XXXXXXX.XXXXXXX}

\acmConference[Preprint '21]{Preprint '21: Preprint}{May 14, 2021}{Washington, DC}
\acmPrice{XX.XX}
\acmISBN{978-1-4503-XXXX-X/21/05}

\usepackage[flushleft]{threeparttable}
\usepackage[caption=false]{subfig}
\graphicspath{ {./figures/} }

\usepackage{multirow}

\begin{document}

\title{Towards Automatic Comparison of Data Privacy Documents: A Preliminary Experiment on GDPR-like Laws}

\author{Kornraphop Kawintiranon}
\authornote{Both authors contributed equally to this research.}
\email{kk1155@georgetown.edu}
\affiliation{%
  \institution{Georgetown University}
  \city{Washington}
  \state{DC}
  \country{USA}
}

\author{Yaguang Liu}
\authornotemark[1]
\email{yl947@georgetown.edu}
\affiliation{%
  \institution{Georgetown University}
  \city{Washington}
  \state{DC}
  \country{USA}
}

\renewcommand{\shortauthors}{Kawintiranon and Liu}

\begin{abstract}
General Data Protection Regulation (GDPR) becomes a standard law for data protection in many countries. Currently, twelve countries adopt the regulation and establish their GDPR-like regulation. However, to evaluate the differences and similarities of these GDPR-like regulations is time-consuming and needs a lot of manual effort from legal experts. Moreover, GDPR-like regulations from different countries are written in their languages leading to a more difficult task since legal experts who know both languages are essential. In this paper, we investigate a simple natural language processing (NLP) approach to tackle the problem. We first extract chunks of information from GDPR-like documents and form structured data from natural language. Next, we use NLP methods to compare documents to measure their similarity. Finally, we manually label a small set of data to evaluate our approach. The empirical result shows that the BERT model with cosine similarity outperforms other baselines. Our data and code are publicly available.\footnote{\url{https://github.com/kornosk/GDPR-similarity-comparison}}
\end{abstract}



\keywords{GDPR, GDPR-like, Personal Data Protection, Data Privacy, Legal Document Analysis, Document Similarity, Natural Language Processing}

\maketitle

\section{Introduction}
\label{sec:introduction}

Internet has become one of the most important things nowadays. The internet has a wide range of advantages such as telecommunication, news, social media and e-commerce. The Pew Research Center reports that over 70\% of American people access internet through their devices~\cite{pew2021more}. When users spend time surfing on the internet, it is possible that your online footprint is tracked by some malware, or the websites intentionally gather your usage information. Commercial organizations could make a lot of benefits from user behavior information. For example, Youtube suggests the new videos that you may like. This recommendation system is built based on artificial intelligence (AI) and machine learning (ML). Apart from the user behavior, sometimes, user information is also collected silently. For example, in order to proceed some activity, a website may ask users for their personal data such as name, age, sex and email. Once the data is collected, users have no access or right to perform any actions about the data. This raises up an issue about privacy and personal data.

To tackle this issue, Europe has created a set of rules to regulate the use of personal data by organizations. This regulation is called \textit{General Data Protection and Regulation (GDPR)}. It becomes the standard for personal data protection law for many countries including Brazil, Australia, United States, Japan, South Korea, Thailand, Chile, New Zealand, India, South Africa, China and Canada.\footnote{There are twelve countries as of April 2021.} The GDPR-like data privacy law from these countries is modified from the original GDPR. However, to measure the similarity and difference among these laws, legal experts are needed. In addition, to fully understand laws in different languages, a complete translation is essential. To assist legal experts in GDPR-like data privacy laws comparison, an automatic approach is necessary.

In this paper, we investigate the use of machine learning (ML) and natural language processing (NLP) to address the issue. Because the law document is written in natural language, we extract the document into a data set where each row is a recital with associated article, section and chapter. The detail of the data transformation is described in~\ref{sec:data_transformation}. Once the data is well structured, we apply NLP techniques for document similarity measurement. As the first pilot test, we chose to experiment on original GDPR and Brazilian law (LGPD) because Brazil also provides an official translated version of LGPD along with the Spanish-language version. We evaluate our approach by manually checking on a small data set and found that the pre-training deep Bidirectional Encoder Representations from Transformers (BERT) outperforms other baselines. However, our labeled data set is very small. We encourage researchers in the field to use our well-structured data for more annotations.

Our contributions are as follows:
\begin{itemize}
   \item To the best of our knowledge, our study is the first to incorporate the NLP techniques to measure the similarity among GDPR-like data privacy laws from different countries.
   \item We extract and transform the content of GDPR-like documents to form a well-structured data set. This could be beneficial for researchers to investigate the use of NLP for these GDPR-like documents.
   \item We provide the article pairs of original GDPR and LGPD with small amount of labels.
   \item We release our data and code to support researchers in the field to study GDPR-like document comparison at~\url{https://github.com/kornosk/GDPR-similarity-comparison}.
\end{itemize}
\section{Background and Related Work}
\label{sec:background_and_related_work}

\subsection{General Data Protection and Regulation (GDPR)}
\label{sec:gdpr}

The General Data Protection Regulation (GDPR) came into effect in 2018 to replace the Data Protection Directive. It was designed to leverage data privacy laws across Europe with the development of new technology and the new challenges and problems in protecting personal privacy 
\cite{EUdataregulations2018}. The structure of GDPR consists of Chapters, Articles, Sections and Recitals. It applies to all the organizations
that handle personal data for EU residents, regardless of
their locations and thus, it is widely viewed as a benchmark
for data protection and privacy regulations and inspired many countries to legislate their own laws to enable greater protection and capabilities for users. For example, Brazil passed the General Data Protection Law (Portuguese: Lei Geral de Proteção de Dados Pessoais, or LGPD) in 2018, and it came into effect in February 2020. India also release their law for user protection, Personal Data Protection Bill (PDPB), which is modeled after the GDPR, though there are significant differences between the Indian and European law. 

\subsection{Legal Document Comparison}
\label{sec:legal_doc_comparison}

To perform a sentence similarity analysis, information retrieval is a key to success since legal documents are generally long. Text summarization is a potential direction for this task since it selects important sentences to represent the entire document. However, summarizing the long document is still a very difficult task. Some research studies~\cite{cachola2020tldr,CAGLIERO2020113659} have proposed a transformer-based and n-gram-based model to summarize long scientific documents but this is limited to use for only scientific papers and incapable of legal document summarization.

Computing the similarity between two legal documents is a very important task in the domain of legal document processing such as information retrieval and legal documents comparisons. This task also has a lot of important downstream applications (e.g. recommendation and ranking). Previous models of legal document similarity mainly focus on two directions. First, regardless of text content, a similarity score can be measured by analyzing the dominant citations network. \citet{kumar2011similarity} proposes a model that is extended to the popular techniques used in information retrieval and search engines using co-citation. Second direction is comparing the content of legal documents using text similarity algorithms. In short, we feed a pair of documents and the algorithm outputs the similarity score indicating how similar they are. A number of researchers exploit the task as an instance of the semantic text matching (STM) problem. \citet{landthaler2018semantic} proposes a purely unsupervised solution for semantic text matching between contract clauses and legal comments. \citet{mandal2017measuring} further develops a model that takes advantage of TF-IDF-based methods and advanced techniques such as word embedding and document embedding. However, the content similarity is still a hard task as understanding the content itself is already challenging.

Many works of literature study case-based instead of document-based where the task is to extract similar or relevant cases “S1, S2, S3, …, Sn” from historical legal documents given the new case “Q”~\cite{vu2019building,nguyen2020jnlp}. In addition to case-based research, a number of studies focus on statute law documents where the task is to extract relevant articles given the question “Q”~\cite{dang2019approach,gain2021iitp,nguyen2020jnlp} and the following task is to answer a given question “Q” by determining if the relevant articles entail "Q" or "not Q"~\cite{hudzina2019statutory,gain2021iitp,nguyen2020jnlp}. These works are closely related to our task but there are some minor differences including \textit{(i)} we compare two documents instead of having an input “Q” then find relevant articles from a corpus of legal documents, \textit{(ii)} the literature mainly focuses on cases or statute laws while we focus on the new type of law and regulation for digital privacy, and \textit{(iii)} many-to-many comparison is a much harder problem than one-to-many because we also need to rank the similar articles of document “D1” for every article in the document “D2”. This means a ranking algorithm~\cite{narayan2018ranking} might be required for this task.

In order to fully exploit the text and better compare similarities, we use BERT~\cite{devlin2019bert}, a pre-trained transformer network, which set for various NLP tasks new state-of-the-art results, including question answering, sentence classification, and sentence-pair regression. Specifically, we convert each sentence into embedding space and then use metrics such as cosine similarity for similarity comparison. Different than previous models which use only word-based approaches, our algorithm is expected to be able to better analyze text and thus gain a better similarity. We will investigate more potential directions along the way. To simplify our study, we focus on English legal documents as the first step. We put the extension to other languages as our future work.
\section{Methodology}
\label{sec:methodology}

There are two steps for GDPR-like document comparison. First, we need to transform the content from natural language to structured data. Next, we process the structured data with similarity algorithms. The detail of each step is described in the following sections.

\subsection{Data Transformation}
\label{sec:data_transformation}

We extract content in natural language and form a well-structured data. The GDPR is comprised of four levels of content. Chapter is the top hierarchy of the law structure, each chapter has many sections. Each section has many articles and there are many recitals in each articles. Note that each recital can include multiple sentences. We extract content into recital-level as it is the smallest piece of information to convey some legal meaning. Because some recitals themselves have no meaning without the relevant contexts, performing document analysis on the upper level (e.g. articles level) would be more informative.

\subsection{Similarity Algorithms}
\label{sec:sim_algo}
In this section, we propose several algorithms from traditional ones to deep learning ones for document comparison. We choose TF-IDF, which is widely used and still one of the most popular ones for legal documents comparison. Then we will introduce deep learning based algorithms.

\subsubsection{TF-IDF}
Simple and widely used method to convert word to numerical data in text mining is TF-IDF. It is a measure that evaluates how relevant a word is to a document in a collection of documents. Two metrics are needed and then multiplied: how many times a word appears in a document, and the inverse document frequency of the word across a set of documents. The formula is defined as below \cite{tfidf}:

\begin{equation}
\label{eq:tf}
    tf(t,d) = \frac{f_{t,d}}{\sum_{t_o}^{} f_{t_o,d}}
    \vspace{2mm}
\end{equation}
\begin{equation}
\label{eq:idf}
    idf(t, D) = log \frac{N}{|\{d \subset D: t \subset d \}|}
    \vspace{5mm}
\end{equation}

From Equation~\ref{eq:tf}, tf(t,d) is the frequency of term t, where $f_{t,d}$ is the raw count of a term in a document. The inverse document frequency computation is shown in Equation~\ref{eq:idf}. idf(t, D) denotes the inverse document frequency where N is the total number of documents in the corpus N = |D|, and $|\{d \subset D: t \subset d \}|$ is the number of documents where the term appears.

The first one explains what words are more important by frequency. However, this could be misleading. For example, the word "the" could appear many times in a document and this will lead the algorithm to emphasize documents which happen to use the word "the" more frequently, without giving enough weight to the more meaningful terms. Thus we need the second metric which is able to filter out words that appear too frequently. Now that we have the representation for each sentence and a document is represented as a matrix since it consists of many sentences. We map the matrix to a vector by summing it. Next, we will use cosine similarity for the documents comparison.

\subsubsection{Word Embedding}
An efficient approach to transform word to a representation vector is word embedding. It is a learned representation for text where words that have the same meaning have a similar representation. For example, "good" and "well" and will be represented as list of numbers that are similar. But "good" and "bad" will have very different vector representations. Here we use GloVe \cite{pennington2014glove} for the embedding space mapping. Specifically, each word is mapped into a vector that consists of a list of numbers using GloVe and the documents of a user are represented as a matrix. We map the matrix features to a vector by summing the matrix. 

\subsubsection{BERT Embedding}
Bidirectional Encoder Representations from Transformers (BERT) was created and published in 2018 from Google and it has become a new standard for Natural Language Processing \cite{devlin2019bert}. It achieved a whole new state-of-the-art on many NLP task, including text classification, sequence labeling, etc. Besides, it works well even with only a small amount of data. Compared with word embedding which analyze text one word at a time, BERT will consider the context. For example, suppose in our data set we have the two sentences: 1) Went to the bank to get money. 2) Went to the river bank to fish. Word embedding will generate the same embedding for the word “bank”. When using sentence embeddings, each sentence gets its own vector representation, therefore, capturing the contextual differences of the two sentences. We again represent document features as a vector by summing the sentence embeddings. 

\subsubsection{Siamese Network Embedding}
We hypothesize that if two sentences are similar, they would have semantically similar embeddings. To capture these clusters, we use a Siamese network architecture to generate sentence embeddings. Specifically, the fine-tuning model is trained on the SNLI \cite{bowman2015large} and the MNLI datasets \cite{williams2017broad}. They are a collection of English sentence pairs manually annotated with the labels entailment, contradiction, and neutral. With the training, similar sentences will have closer embeddings while dissimilar ones will have more different embeddings. With the new model, we are able to get the textual embeddings and the rest if the same as Word Embedding model and  BERT Sentence Embedding model.
\section{Experiments}
\label{sec:experiments}

\subsection{Experimental Settings}
\label{sec:experimental_settings}
\begin{figure*}[!htb]
    \centering
    \includegraphics[width=0.8\textwidth]{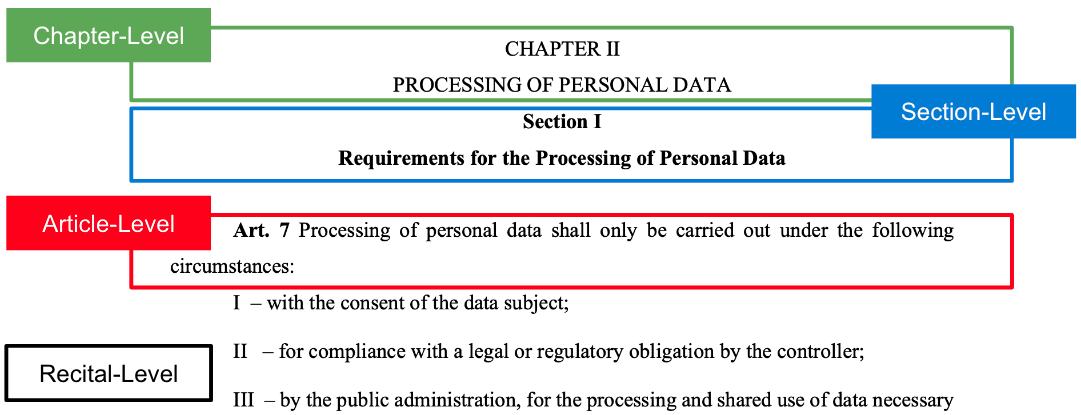}
    \caption{The example of LGPD document structures.}
    \label{fig:example-structure}
\end{figure*}

 We choose the Brazilian GDPR-like law (LGPD) to compare to the original GDPR (European version). There are two reasons for this. First, the originally, LGPD was written in Spanish then translated into English version. This helps us by short-cutting the translation step from non-English to English language in order to compare with GDPR. Second, we investigated LGPD and found that it has the same document structure as GDPR including Chapter, Section, Article and Recital. The example of LGPD is shown in Figure~\ref{fig:example-structure}.
 
\subsection{Evaluation Methods}
\label{sec:evaluation_methods}

To evaluate system in the document similarity ranking, we use HIT@K evaluation method. It would determine the correct match only if among the top K similar document retrieved from corpus, there is at least one document is actually the right match. Formally, given an input document $d^a$ from corpus $C_a$, we get $K$ most similar documents $d^b_1$, $d^b_2$, ..., $d^b_K$ retrieved from the other corpus $C_b$. Equation~\ref{eq:hitk-accuracy} shows the accuracy as average of corrects where $n$ is the total number of documents in $C_a$ that we evaluate. The prediction is determined as correct if at least one of top K similar documents $d^b_j$ where $j \in [1,K]$ is matched to the input $d^a$.

\begin{equation}
\label{eq:hitk-is-match}
    is\_match(d_i, d_j) =
    \begin{cases}
        1,              & \text{if } d_i \text{ is related to } d_j\\
        0,              & \text{otherwise}
    \end{cases}
    \vspace{2mm}
\end{equation}

\begin{equation}
\label{eq:hitk-accuracy}
    Accuracy = \frac{\sum_{i=1}^{i=|D^a|}\sum_{j=1}^{j=K} is\_match(d^a_i, d^b_j)}{|D^a|}
    \vspace{5mm}
\end{equation}

\begin{figure*}[!htb]
    \centering
    \subfloat[Correct Match]{%
        \includegraphics[width=\textwidth]{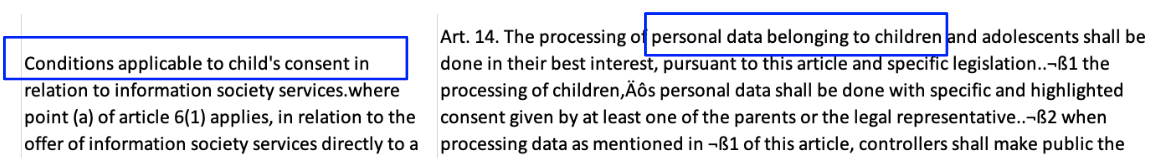}%
    }
    \vspace{-2mm}
    \subfloat[Incorrect Match]{%
        \includegraphics[width=\textwidth]{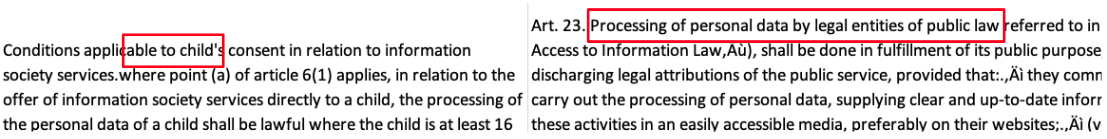}%
    }
    \caption{Examples of correct and incorrect match between articles from GDPR and LGPD.}
    \label{fig:example-match}
\end{figure*}

In this paper, we choose HIT@1 (we set $K=1$) as our evaluation metric. We conduct comparison experiment in article-level and recital-level because we think that directly using similarity algorithms would not be able to handle very long documents. The example of correct match and incorrect match in article-level are shown in Figure~\ref{fig:example-match}.

There are over 80 articles in GDPR, so we choose to evaluate 10 articles as our pilot test. These articles are from different topics including Subject matters, Material scope, Territorial scope, Definition of personal data, Principles of personal data usage, consent, Conditions for consent, Child's consent, Special types of personal data and Personal data relating to criminal. Two researchers manually check all these articles and their matches in order to determine the corrects. If we have disagreement, we discuss and finalize the agreed labels. The accuracy is the average corrects of these ten manually labeled article pairs.

\subsection{Experimental Results}
\label{sec:experimental_results}

We report our preliminary results based on experimental setup and evaluation methods described in Section~\ref{sec:experimental_settings} and ~\ref{sec:evaluation_methods}, respectively. Table \ref{table:accuracy} shows the result of accuracy for GDPR and Brazilian GDPR (LGPD). Both experimental results on recital and article level are included.

\begin{table}[!htb]
\begin{tabular}{|c|c|c|}
\hline
Level                    & Algorithm      & HIT@1 \\ \hline
\multirow{4}{*}{Recital} & Siamese BERT   & 0.6   \\ \cline{2-3} 
                         & TF-IDF         & 0.6   \\ \cline{2-3} 
                         & Word Embedding & 0.2   \\ \cline{2-3} 
                         & BERT           & 0.7   \\ \hline
\multirow{4}{*}{Article} & Siamese BERT   & 0.4   \\ \cline{2-3} 
                         & TF-IDF         & 0.5   \\ \cline{2-3} 
                         & Word Embedding & 0.5   \\ \cline{2-3} 
                         & BERT           & 0.7   \\ \hline
\end{tabular}
\caption{GDPR vs. Brazilian GDPR (LGPD) accuracy for Recital level
 and Article level}
\label{table:accuracy}
\end{table}

For recital level, the result from the models could vary for different topics since we are using only one or a few sentences. It could be affected by the noise of texts, the length of the recitals, etc. But it still depicts the effects of different models. BERT gives the best result in terms of accuracy. This matches the recent NLP papers that pretrained model on big data using deep learning technique can provide a better sense and representation of texts. Word embedding performs the worst. We think this is because 1) GloVe, the model we use for generating embedding space is relatively old 2) word embedding is trained on general texts such as wikipedia, etc, it probably can not improve legal documents. However, more sentences can better reflect the model. We will analyze this in the next section. The result for TF-IDF is similar to other papers where models often rely on this feature to find related or similar documents. Siamese embedding performs the second best, as TF-IDF, although the result is supposed to be better since this model is trained on gold label data. The reason is because that Siamese BERT is trained on pairs of sentences. For these sentences, words such as “should” and “must” will have a very similar representation. This makes general texts more understandable but for law documents, this can get inaccurate.

For article level, the effect, no matter a good effect or a bad one, from the models will be accumulated and thus the results will reflect the algorithm quality better. BERT gives the best result. This matches the recital level and further confirm that recent NLP developments with deep learning technique can help the legal document tasks. Word embedding and TF-IDF are similar. We think that since word embedding is trained on general texts, it should just be a reflection of the general word, just like TF-IDF using word count and similar techniques. Siamese embedding performs the worst. With more sentences, the siamese network trained on the general documents is not able to properly recognize the differences between different documents. The scores among different models are more smooth and thus further confirm more data could lead to a more stable result.

\section{Conclusions and future direction}
\label{sec:conclusions}
In this paper, we develop an end-to-end automatic approach for GDPR-like documents similarity checking. We apply state-of-the-art models from NLP community to our experiments and show that they outperform the TF-IDF method which is commonly used in legal documents comparison. However, the evaluation is a pilot test and it only shows the baseline results for the task. 

In the future, more sophisticated study specifically for GDPR similarity checking is needed and a number of cutting-edge algorithms could be investigated. For example, We could take into account other GDPR-like languages (e.g. Chinese or Thai version). Besides, better evaluation of performance should be developed. Although we try our best, neither of us are experts in the area so the result may be not absolutely correct. In this paper, we only consider HIT@1. Similarity scores could also be taken into account to reflect better on the documents comparison instead of considering only search ranking. Finally, we release code and data sets in order for researchers who would like to study GDPR-like document comparison.

\section*{Acknowledgements}
This research is part of our Ph.D. seminar at Georgetown University (COSC 824 - Data Protection by Design). We would like to thank our professor, Benjamin Ujcich, and the members of the seminar for their comments and suggestions. In the midst of pandemic, we would also like to thank our family who always support us.

\bibliographystyle{ACM-Reference-Format}
\bibliography{ours}










\end{document}